# EXPERT OPERATIONAL GANS: TOWARDS REAL-COLOR UNDERWATER IMAGE RESTORATION

*Ozer Can Devecioglu[1], Serkan Kiranyaz[2], Mehmet Yamac[1], and Moncef Gabbouj[1]*

[1]Department of Computing Sciences, Tampere University, Tampere, Finland.
[2]Electrical Engineering Department, Qatar University, Doha, Qatar.

## ABSTRACT

The wide range of deformation artifacts that arise from complex light propagation, scattering, and depth-dependent attenuation makes the underwater image restoration to remain a challenging problem. Like other single deep regressor networks, conventional GAN-based restoration methods struggle to perform well across this heterogeneous domain, since a single generator network is typically insufficient to capture the full range of visual degradations. In order to overcome this limitation, we propose xOp-GAN, a novel GAN model with several expert generator networks, each trained solely on a particular subset with a certain image quality. Thus, each generator can learn to maximize its restoration performance for a particular quality range. Once a xOp-GAN is trained, each generator can restore the input image and the best restored image can then be selected by the discriminator based on its perceptual confidence score. As a result, xOP-GAN is the first GAN model with multiple generators where the discriminator is being used during the inference of the regression task. Experimental results on benchmark Large Scale Underwater Image (LSUI) dataset demonstrates that xOp-GAN achieves PSNR levels up to 25.16 dB, surpassing all single-regressor models by a large margin even, with reduced complexity.

## 1. INTRODUCTION

Underwater image restoration is a very challenging image processing problem due to the presence of numerous types of visual abnormalities of different severity. These artifacts usually involve light scattering, blur, color distortion, and low contrast. Figure 1 illustrates examples from the LSUI dataset, the largest benchmark dataset ever composed, showing both corrupted images and their corresponding ground truths. The corrupted images are uniformly partitioned based on their PSNR (Peak Signal-to-Noise Ratio) values into three categories: low (8.79–16.57 dB), medium (16.57–19.79 dB), and high quality (19.79–31.53 dB).

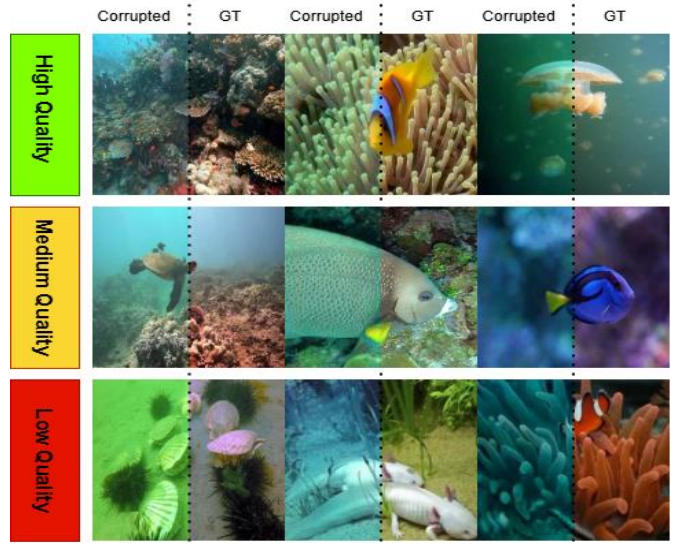

Figure 1: Sample corrupted images (left) with varying PSNR levels from the LSUI dataset with their ground-truth (target) images (right).

This categorization reflects the wide range of artifact severity levels commonly encountered in underwater environments. To effectively restore such images, various degradation types must be addressed over the same image, which remains a challenging problem even with deep and fine-tuned regressor networks proposed in the literature. Among them, the highest PSNR performance on the LSUI dataset was achieved by Peng et al. [1], who used a U-Shape Transformer with over 65M parameters, reaching 24.16 dB PSNR level, which sets the current SoTA performance level. Many recent prior works could not even surpass 20dB PSNR. For example Chiang et al. [2] proposed a wavelength compensation and dehazing method, reporting 19.72 dB and 18.41 dB PSNR at 5 m and 15 m depths, respectively, using a small video-derived dataset [3]. Luo et al. [6] focused on color correction and contrast enhancement, reporting 12 dB PSNR on 18 images. Awan et al. [7] introduced UW-Net using wavelet transforms, achieving around 19.9 dB on the LSUI dataset.

Transformer-based models have recently gained momentum in the field of image restoration, including underwater improvement, thanks to their ability to model long-range dependencies and global context. While these models have usually produced superior outcomes, they come along with a massive computational cost and huge data requirements for training, making them not appropriate for real-time or resource-constrained applications. Generative Adversarial Networks (GANs) [8], on the other hand, continue to demonstrate outstanding potential for generating perceptually convincing and structurally consistent restorations. However, despite such advantages, even their learning performance can be limited due totheir intrinsic complexity and significant variability in the quality ofunderwater images. As observed in Figure 1, the severity of artifacts and color bias can vary substantially between samples, making it challenging for a single regressor (i.e., the GAN generator) to effectively restore input images with varying levels of degradation. To fully realize the full potential of GANs in underwater image restoration and possibly other challenging regression tasks, we proposeto make the learning process more effective by adopting a "Divide & Conquer" type approach with multiple generators. In this paper, we introduce a novel GAN model that overcomes this problem with multiple expert generators each of which is trained over severity-aware data partitions. The proposed model is, therefore, called E**x**pert **Op**erational **GANs** (xOp-GANs), which consists of three specialized generators and a discriminator. Unlike the standard GAN models with a single generator, each generator in an xOp-GAN is trained to specialize on a specific data partition corresponding to a certain level of visual degradation. This customized training enables each generator to develop *expertise* in restoring images within a certain quality range, thus enhancing overall restoration fidelity. Unlike the traditional GANs, during the inference, all three generators independently restore the input image and then the discriminator is used to select the best restored image (based on the perceptual quality of each output) using its confidence score. To further boost the learning performance, instead of conventional convolutional layers, all generator and discriminator networks are redesigned using Self Organized Operational Neural Network (Self-ONN) layers [9]-[11].

The rest of this article is organized as follows: in Section 2 a brief introduction to Self-ONNs will first be presented. Then the proposed xOp-GANs will be detailed. In Section 3, the benchmark LSUI dataset is introduced, and an extensive set of comparative evaluations will be presented over the LSUI dataset. Conclusive remarks will be drawn and topics for future research will be discussed in Section 4.

## 2. METHODOLOGY

### 2.1. Self-Operational Neural Networks

In this section, we provide a concise overview of Self-ONNs and highlight their fundamental properties. Unlike the fixed convolutional operations used in traditional CNNs, the generative neurons in a Self-ONN utilize nodal operators capable of performing flexible nonlinear transformations. These transformations can be represented using a Taylor series expansion around a point close to the origin.

$$\psi(x) = \sum_{n=0}^{\infty} \frac{\psi^{(n)}(0)}{n!} x^n \quad (1)$$

This formulation allows for the approximation of any arbitrary function ψ near zero. When activation functions such as the hyperbolic tangent (tanh) constrain the input feature maps of the neurons within a region close to zero, the expression in Equation (1) can be leveraged to construct a composite nodal operator. By denoting $\frac{\psi^{(n)}(0)}{n!}$ as the $w_n$ , the $n^{th}$ parameter in Q-dimensional Taylor approximation, , the coefficients can be compactly represented as follows:

$$\psi(w, y) = w_0 + w_1 y + w_2 y^2 + \cdots + w_Q y^Q \quad (2)$$

In this context, $w_0$ represents the bias term, while $w_1$ through $w_Q$ correspond to the Maclaurin series coefficients, all of which are learned during backpropagation. For a more detailed explanation of the theoretical foundation and forward propagation mechanism of Self-ONNs, readers are referred to [9]-[11].

### 2.2. Expert Operational GANs (xOp-GANs)

Underwater images exhibit a wide variety of visual degradation due to the complexity of environmental factors such as light absorption, scattering, and suspended particles. This variation makes it exceedingly difficult for a single restoration network to generalize effectively across the full spectrum of distortions. Instead of increasing network depth or capacity, we propose xOp-GAN, a multi-branch generative framework composed of three expert generators, each dedicated to learning restorations within a specific quality range. The overall architecture of the proposed xOp-GAN is illustrated in Figure 2. In the training step, the dataset is first grouped into three subsets based on PSNR levels: High Quality (HQ), Medium Quality (MQ), and Low Quality (LQ) images. Prior to training, all images are normalized to the range between -1 and 1. Each generator in the xOp-GAN is then exclusively trained on one of these subsets using a loss function tailored to the GAN training objective, enabling the generators to specialize in recovering images exhibiting artifacts characteristic to their assigned quality class. The conventional objective functions for Op. GANs as expressed in Eq. (3) are used as the loss functions during training.

$$\sum_{i \in \{HQ, MQ, LQ\}} \min_{OG} \max_{OD} L_{OP-GAN}(OG_i, OD)$$
$$= E[\log(OD(GT_i))]$$
$$+ E[\log(1 - OD(OG_i(X_i)))] \quad (3)$$

where, $OG_i$ and $OD$ represent the Operational Generator and Operational Discriminator, respectively. $X_i$ denotes the input, and $GT_i$ stands for the corresponding ground truth. During adversarial training, the discriminator is exposed not only to the outputs of the corresponding generator but also to the outputs of the other generators to simulate a realistic

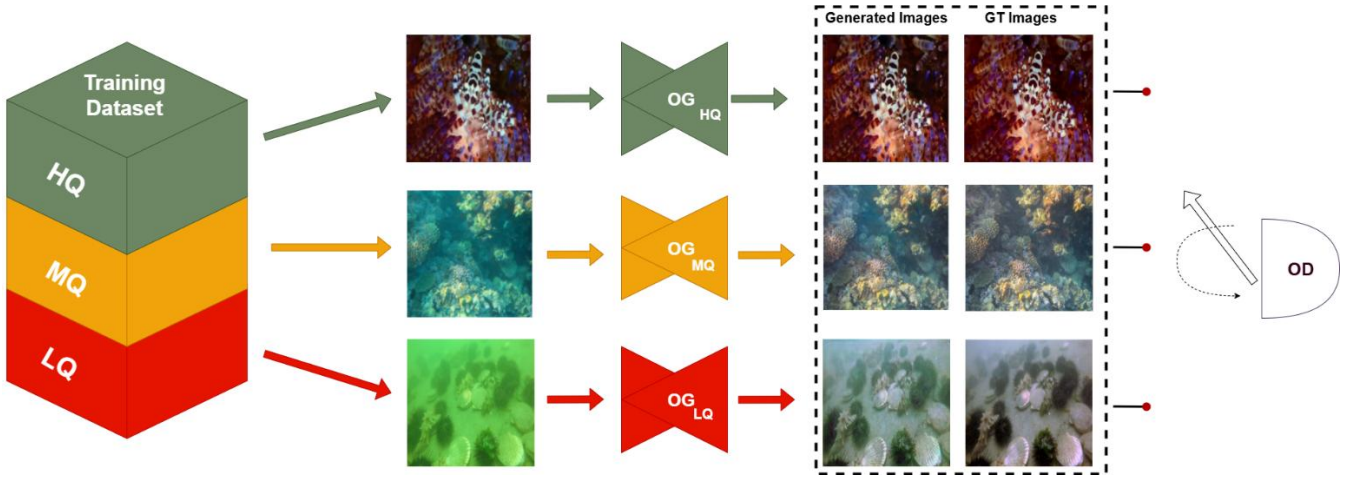

Figure 2: The general training framework for xOp-GANs with Operational Generators (**OGs**) and Discriminator (**OD**).

inference scenario where the quality level of an input image is unknown.

To prevent overfitting and promote more robust learning, the discriminator is trained with instead of binary labels, random values in the range [0, 2$\epsilon$] are used for fake samples, and values in [1-2$\epsilon$, 1] for real samples where $\epsilon$ is a practical range parameter:

$$\sum_{i\in\{HQ,MQ,LQ\}} \sum_{k\in\{HQ,MQ,LQ\}} \|OD(OG_i(X_k)), \epsilon \mp \epsilon\| + \|GT_k, 0.9 \mp \epsilon\| \quad (4)$$

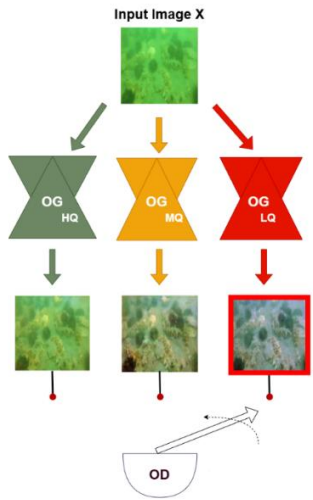

Figure 3: The inference of the xOp-GAN model with three OGs. Each OG output is evaluated by the OD and the best output is selected with respect to their confidence score.

During the restoration of any test image, in contrast to traditional GAN frameworks where the discriminator is discarded, the discriminator remains active and plays a crucial role in the xOp-GANs' inference process. As illustrated in Figure 3, given an unseen input image **X**, all three generators independently create their restored outputs. These outputs are then evaluated by the discriminator, and the image corresponding to the highest confidence score is selected as the (best) restored output. This design enables xOp-GAN to dynamically adapt to unknown input quality levels without explicit classification, turning the discriminator into a perceptual assessment tool during the inference phase.

## 3. EXPERIMENTAL RESULTS

### 3.1. Large Scale Underwater Image Dataset

LSUI is the largest benchmark underwater image dataset including 4279 image pairs, which covers abundant real\underwater scenes with high-quality ground-truth (GT) images. The dataset also contains underwater images from other benchmark underwater datasets. The ground-truth (GT) underwater images are manually selected using a 2-step evaluation scheme. First, they used an ensemble model that used 18 existing underwater image enhancement methods [1]. After the first round, volunteers and experts voted for the best-enhanced image to form the paired dataset. The distribution of PSNR values in the training set reveals a wide range of degradation levels, spanning from 8.79 dB to 31.53 dB. For fair comparative evaluations with the prior works including the SoTA model in [1], the same train-test partitioning (90%-10%) is applied on the LSUI dataset. For xOp-GAN training, the train set are divided into three equal-size partitions based on their PSNR values: low quality (LQ:8.79–16.57 dB), medium quality (MQ: 16.57–19.79 dB), and high quality (HQ: 19.79–31.53 dB). Each partition contains 1280 input-target (original-GT) image pairs.

### 3.2. Experimental Setup

For the Operational Generator (OG) models, a 10-layer U-Net architecture is employed. The network comprises 5

operational layers in the encoder and 5 upsampling blocks in the decoder, where each upsampling block also includes operational layers with residual connections instead of transposed convolutions. In the encoder part of the OG, all operational layers use a kernel size of 7 and a stride of 2. In the decoder, the operational layers have a kernel size of 5, with standard upsampling by a factor of 2. The nonlinearity degree (Q value) is set to 3 for all layers within the generators.

The Operational Discriminator (OD) consists of 5 operational layers followed by 2 fully connected (dense) layers. The operational layers have a kernel size of 4, with stride values of 4, 4, 4, 2, and 2, respectively. The Q value for all discriminator layers is set to 2.

All OGs are initialized using pre-trained weights obtained by training an Op-GAN on the entire dataset, prior to specialization. For all experiments, the batch size is set to 1, and training is conducted for a maximum of 5000 backpropagation (BP) iterations using the Adam optimizer. The initial learning rate is set to $10^{-5}$. We implemented both OG and OD networks using the FastONN library [12] based on PyTorch.

### 3.3. Results

We evaluated the performance of the proposed xOp-GAN model on the LSUI dataset and compared it with several state-of-the-art underwater image restoration methods in terms of PSNR and model complexity (parameter count in millions). As shown in Table 1, when the best generator output is selected using ground-truth PSNR, xOp-GAN achieves a score of 25.16 dB, surpassing all baseline models and setting the new SoTA level in underwater image restoration. When the selection is made by the discriminator during inference, the model still reaches 23.44 dB, making it the second-best performing model after the U-Shape Transformer (24.16 dB), despite using nearly three times fewer parameters (24.9M vs. 65.6M). Most importantly, both results indicate that xOp-GANs can outperform traditional (baseline) Op-GANs with a significant performance gap. Moreover, it outperforms other deep and learning-based methods such as UColor, UGAN, and WaterNet with a large margin proving its efficiency and restoration effectiveness. These results validate that xOp-GAN provides a strong balance between accuracy and computational cost, making it a compelling solution for real-time or resource-constrained underwater imaging scenarios.

Figure 4 shows the visual comparison of generator outputs for different input quality levels. Each input image is passed through all three OGs. The red dashed boxes indicate the output selected by the discriminator during inference based on the highest confidence score. The three generators produce noticeably different outputs, each tailored to specific types of corruption artifacts. As highlighted in the figure, the discriminator may fail to select the output with the highest PSNR level but the color distribution of the selected output even with a lower PSNR usually seems to be more realistic even than the GT. This is possible since GT images were selected manually among a group of outputs from different methods. Therefore, restoring "even better" outputs is possible but of course with the reduced PSNR levels. This further demonstrates the discriminator's ability to capture perceptual quality better than traditional metrics and supports more reliable restoration for underwater images toward their true-color appearance.

**Table 1**: The underwater image restoration performance over LSUI Dataset. * indicates the best output performance of the xOp-GANs.

|  | PSNR (dB) | Parameters (M) |
|---|---|---|
| U-Shape Transformer [1] | 24.16 | 65.6 |
| xOp. GANs* | **25.16** | 22.8 |
| xOp.GANs | 23.45 | 24.9 |
| Op. GANs [14] | 22.93 | 7.6 |
| UColor [15] | 22.91 | 157.4 |
| UIE-DAL [16] | 17.45 | 18.82 |
| UGAN [17] | 19.79 | 57.17 |
| FunIE [18] | 19.79 | **7.01** |
| WaterNet [19] | 17.73 | 24.81 |

## 4. CONCLUSION

In this study, we introduced the xOp-GAN model, a novel multi-generator GAN framework developed for robust underwater image restoration in the context of severe and diverse visual degradations. xOp-GAN uses three expert Generators, each specialized in the restoration of images in certain quality ranges. With the proposed training approach, xOp-GANs can improve domain representation and restoration quality across a wide-range of degradation levels. During the inference, its discriminator can now be used to select the best restoration output. This allows the discriminator to actively participate in the restoration rather than being used exclusively during training only, which is common for traditional GAN models.

Experimental results demonstrate that xOp-GANs can significantly outperform the operational GAN model under fair conditions. When the best generator output is selected using ground-truth PSNR, xOp-GAN achieves the highest PSNR level of 25.16 dB, outperforming all baseline models including the SoTA Transformer model with a 1dB margin. This clearly shows its potential and superiority on such a challenging regression problem and promises a powerful alternative for other similar regression problems. Even when relying solely on discriminator-guided selection during inference, the model reaches 23.44 dB PSNR level making it the second best model. This also highlights the fact that there is still room for improvement on the evaluation accuracy of the discriminator during inference. This will be the topic of our future research work.

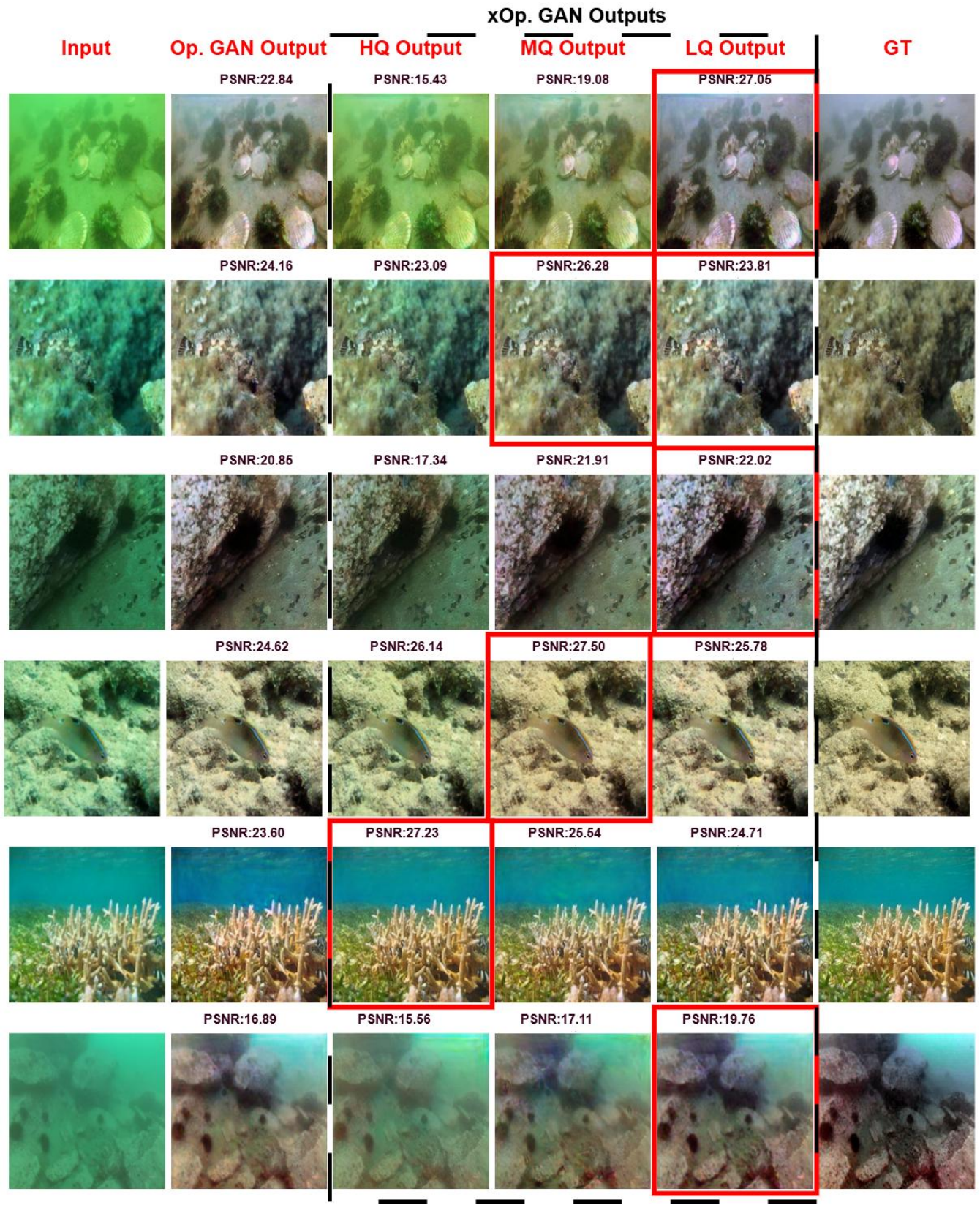

Figure 4: Restoration results from the Op. GAN, (2nd column), OG outputs (2nd-4th columns) of the xOp-GAN, and the corresponding Ground Truth (GT). The OD selected results are highlighted with the red rectangle.

## REFERENCES

[1] L. Peng, C. Zhu and L. Bian, "U-Shape Transformer for Underwater Image Enhancement," in IEEE Transactions on Image Processing, vol. 32, pp. 3066-3079, 2023, doi: 10.1109/TIP.2023.3276332.

[2] J. Y. Chiang and Y. -C. Chen, "Underwater Image Enhancement by Wavelength Compensation and Dehazing," in IEEE Transactions on Image Processing, vol. 21, no. 4, pp. 1756-1769, April 2012, doi: 10.1109/TIP.2011.2179666.

[3] Underwater Image Enhancement by Wavelength Compensation and Dehazing," in IEEE Transactions on Image Processing

[4] P. Liu, G. Wang, H. Qi, C. Zhang, H. Zheng and Z. Yu, "Underwater Image Enhancement With a Deep Residual Framework," in IEEE Access, vol. 7, pp. 94614-94629, 2019, doi: 10.1109/ACCESS.2019.2928976.

[5] K. He, X. Zhang, S. Ren and J. Sun, "Deep Residual Learning for Image Recognition," 2016 IEEE Conference on Computer Vision and Pattern Recognition (CVPR), Las Vegas, NV, USA, 2016, pp. 770-778, doi: 10.1109/CVPR.2016.90.

[6] W. Luo, S. Duan and J. Zheng, "Underwater Image Restoration and Enhancement Based on a Fusion Algorithm With Color Balance, Contrast Optimization, and Histogram Stretching," in IEEE Access, vol. 9, pp. 31792-31804, 2021, doi: 10.1109/ACCESS.2021.3060947.

[7] Awan, Hafiz Shakeel Ahmad, and Muhammad Tariq Mahmood. 2024. "Underwater Image Restoration through Color Correction and UW-Net" Electronics 13, no. 1: 199. https://doi.org/10.3390/electronics13010199

[8] Goodfellow I, Pouget-Abadie J, Mirza M, Xu B, Warde-Farley D, Ozair S, et al. Generative adversarial nets. In: Advances in neural information processing systems. 2014. p. 2672–80.

[9] S. Kiranyaz, J. Malik, H. B. Abdallah, T. Ince, A. Iosifidis, M. Gabbouj, "Self-Organized Operational Neural Networks with Generative Neurons", IEEE Trans. Of Neural Networks and Learning Systems, (Under Review) arXiv preprint arXiv:2004.11778, 2020.

[10] S. Kiranyaz, J. Malik, H. B. Abdallah, T. Ince, A. Iosifidis, and M. Gabbouj, "Self-Organized Operational Neural Networks with Generative Neurons," Neural Networks, vol. 140, pp. 294-308, 2021, doi: 10.1016/j.neunet.2021.02.028.

[11] J. Malik, S. Kiranyaz, and M. Gabbouj, "Self-Organized Operational Neural Networks for Severe Image Restoration Problems," Neural Networks, vol. 135, pp. 201-211, 2021, doi: 10.1016/j.neunet.20

[12] J. Malik, S. Kiranyaz, and M. Gabbouj. (2020). FastONN-- Python based open-source GPU implementation for Operational Neural Networks. arXiv. [Online]. Available: https://arxiv.org/abs/2006.02267

[13] U. Phan, O. C. Devecioglu, S: Kiranyaz, M. Gabbouj, (2025), Progressive Transfer Learning for Multi-Pass Fundus Image Restoration, arXiv preprint arXiv:2504.10025v1

[14] S. Kiranyaz, O.C. Devecioglu, A. Alhams, S. Sassi, T. Ince, O. Abdeljaber, O. Avci, and M. Gabbouj, "Zero-shot motor health monitoring by blind domain transition," Mechanical Systems and Signal Processing, vol. 210, 2024, pp. 111147, doi: 10.1016/j.ymssp.2024.111147

[15] C. Li, S. Anwar, J. Hou, R. Cong, C. Guo, and W. Ren, "Underwater image enhancement via medium transmission-guided multi-color space embedding," IEEE T. Image Process., vol. 30, pp. 4985–5000, 2021.

[16] P. M. Uplavikar, Z. Wu, and Z. Wang, "All-in-one underwater image enhancement using domain-adversarial learning." in CVPR Workshops, 2019, pp. 1–8.

[17] C. Fabbri, M. J. Islam, and J. Sattar, "Enhancing underwater imagery using generative adversarial networks," ICRA, pp. 7159–7165, 2018.

[18] M. J. Islam, Y. Xia, and J. Sattar, "Fast underwater image enhancement for improved visual perception," IEEE Robot. Autom. Lett., vol. 5, no. 2, pp. 3227–3234, 2020

[19] C. Li, C. Guo, W. Ren, R. Cong, J. Hou, S. Kwong, and D. Tao, "An underwater image enhancement benchmark dataset and beyond," IEEE T. Image Process., vol. 29, pp. 4376–4389, 2020.